\pdfoutput=1

\documentclass[11pt]{article}

\usepackage{afterpage}
\usepackage{authblk}
\usepackage[]{ACL2023}
\usepackage{times}
\usepackage{latexsym}

\usepackage[T1]{fontenc}

\usepackage[utf8]{inputenc}

\usepackage{microtype}

\usepackage{inconsolata}
\usepackage{graphicx}
\usepackage{multirow}

\title{CFBenchmark: Chinese Financial Assistant Benchmark \\ for Large Language Model}

\author[1]{Yang Lei}
\author[1]{Jiangtong Li}
\author[1,2]{Dawei Cheng}
\author[1,2]{Zhijun Ding}
\author[1,2]{Changjun Jiang}
\affil[1]{Department of Computer Science and Technology, Tongji University, Shanghai, China.}
\affil[2]{Shanghai Artificial Intelligence Laboratory, Shanghai, China.}

\begin{document}
\maketitle
\begin{abstract}
Large language models (LLMs) have demonstrated great potential in the financial domain. Thus, it becomes important to assess the performance of LLMs in the financial tasks.
In this work, we introduce CFBenchmark, to evaluate the performance of LLMs for Chinese financial assistant.
The basic version of CFBenchmark is designed to evaluate the basic ability in Chinese financial text processing from three aspects~(\emph{i.e.} recognition, classification, and generation) including eight tasks, and includes financial texts ranging in length from 50 to over 1,800 characters.
We conduct experiments on several LLMs available in the literature with CFBenchmark-Basic, and the experimental results indicate that while some LLMs show outstanding performance in specific tasks, overall, there is still significant room for improvement in basic tasks of financial text processing with existing models.
In the future, we plan to explore the advanced version of CFBenchmark, aiming to further explore the extensive capabilities of language models in more profound dimensions as a financial assistant in Chinese. Our codes are released at \url{https://github.com/TongjiFinLab/CFBenchmark}\footnote{Correspondence to dcheng@tongji.edu.cn}.
\end{abstract}

\section{Introduction}

In recent years, with the rapid development of Large Language Models~(LLMs), outstanding performance has been achieved in various tasks by existing LLMs, \emph{e.g.}, ChatGPT~\cite{chatgpt}, GPT-4~\cite{gpt4}, ERNIE~\cite{sun2021ernie}, LLaMA~\cite{touvron2023llama,touvron2023llama2}, ChatGLM~\cite{du2021glm,zeng2022glm}, InternLM~\cite{team2023internlm}, and Baichuan~\cite{yang2023baichuan}.
At the same time, evaluation systems for LLMs have also rapidly evolved, and a series of benchmarks for evaluating these large models have been proposed, \emph{e.g.} GAOKAO-Bench~\cite{zhang2023evaluating}, C-Eval~\cite{huang2023c}, HELM~\cite{liang2022holistic}, and MMLU~\cite{hendrycks2020measuring}, BIG-bench~\cite{srivastava2022beyond}.
However, we notice that there is currently a limited amount of benchmarks focused on assessing the performance of LLMs in specific domains.

Finance is a significant domain that plays an important role in human society \cite{shiller2013finance}.
However, the selection of the finance domain as the focus of our benchmark test is not primarily due to this well-established importance.
Financial domain issues often span various sectors within society, requiring decision-makers to read, compute, analyze, and make decisions based on vast amounts of text and numerical data \cite{subramanyam2014financial}.
Thus, addressing financial problems necessitates that decision-makers possess robust reading, comprehension, and analytical abilities ~\cite{financialcapabilities2005}.
Furthermore, they must have the capacity to integrate knowledge effectively, ultimately leading to precise insights and outstanding expressive skills.
Therefore, a benchmark system tailored for the financial domain concerning LLMs needs to cover the following tasks, ranging from basic to advanced: financial text processing, financial knowledge acquisition, financial calculations, and financial compositional reasoning.
Given the distinctive competency requirements associated with handling financial issues, we have chosen the Chinese finance domain to construct our benchmark system.

In this paper, we introduce the CFBenchmark, a Chinese financial assistant benchmark for large language model. The basic version of CFBenchmark is designed to evaluate the basic ability of LLMs in Chinese financial text processing. It includes {3917} financial texts and is organized around three aspects: financial recognition, financial classification, and financial generation, spanning eight tasks.
For each evaluation aspect, we design several specific tasks to provide a comprehensive evaluation of the Chinese financial text processing ability.
Specifically, for financial recognition, we explore company recognition and product recognition; for financial classification, we explore sentiment analysis, sector classification, and event detection; for financial generation, we explore the content summary, investment suggestion, and risk alert.

Furthermore, we conduct experiments in zero-shot and few-shot mode on our CFBenchmark-Basic to evaluate renowned LLMs available in the literture.
Experimental results reveal that some LLMs show outstanding performance in specific tasks, \emph{e.g.}, ERNIE-Bot-4 achieves 0.618 in financial recognition, and InternLM-20B achieves 0.695 in financial generation.
However, in a general viewpoint, there is no publicly available method that achieves 0.6 in all three tasks, which indicates that there is still significant room for improvement in basic tasks of financial text processing with existing LLMs.
Moreover, we further evaluate the CFGPT1-sft-7B-LoRA and CFGPT1-sft-7B-Full in our CFBenchmark-Basic and achieve better results, which further indicates the effectiveness of domain-specific LLMs.

Our contribution can be summarized as
\begin{itemize}
    \item We analyze the requirements of LLMs in the financial domain, based on which we provide the roadmap for a benchmark system LLMs in the Chinese financial domain.
    We introduce the CFBenchmark, a Chinese financial assistant benchmark for large language model. The basic version of CFBenchmark covers three aspects and eight tasks with a total of {3917} financial texts ranging in length from 50 to over 1,800 characters.
    \item We conduct experiments on the CFBenchmark to evaluate 22 renowned LLMs available in the literature and reveal that there is still a significant room to improve LLMs in basic tasks of financial text processing.
\end{itemize}

\section{Related Work}
In this section, we discuss the related work associated with our benchmark system.

Some domain-focused benchmarks have been introduced to assess the efficiency of diverse approaches in representing text in the financial domain.
For example, FLUE~\cite{shah2022flue} focuses on the English financial domain, whose tasks are designed to testify the classification abilities~(\emph{e.g.}, sentiment, news headlines, and named entities), boundary detection abilities, and financial-related question answering.
Building upon FLUE, FLARE~\cite{xie2023pixiu} added the evaluation of time-series processing capabilities, \emph{i.e.}, forecasting stock price movements.
The FinCUGE~\cite{lu2023bbt} focuses the short financial text, including, sentiment analysis, summary generation, company relation extraction, question answering, and negative subject identification in social media and short news.
FinEval~\cite{zhang2023fineval} is proposed to assess the financial knowledge of LLMs through multiple-choice question-answering including finance, economy and accounting.

These financial domain benchmarks provide valuable references and inspiration for our test construction. However, there are some remaining limitations to be addressed, such as financial event detection, risk summarization, etc.
In this work, we analyze the requirements of LLMs in the financial domain and provide a roadmap for an LLM benchmark system for the Chinese financial assistant.
Then, we introduce CFBenchmark-Basic, the basic version of the Chinese financial assistant benchmark (CFBenchmark), to evaluate the financial text processing capabilities of LLMs with text lengths ranging from 50 to over 1,800 characters.

\begin{figure*}[t]
  \centering
   \includegraphics[width=\textwidth]{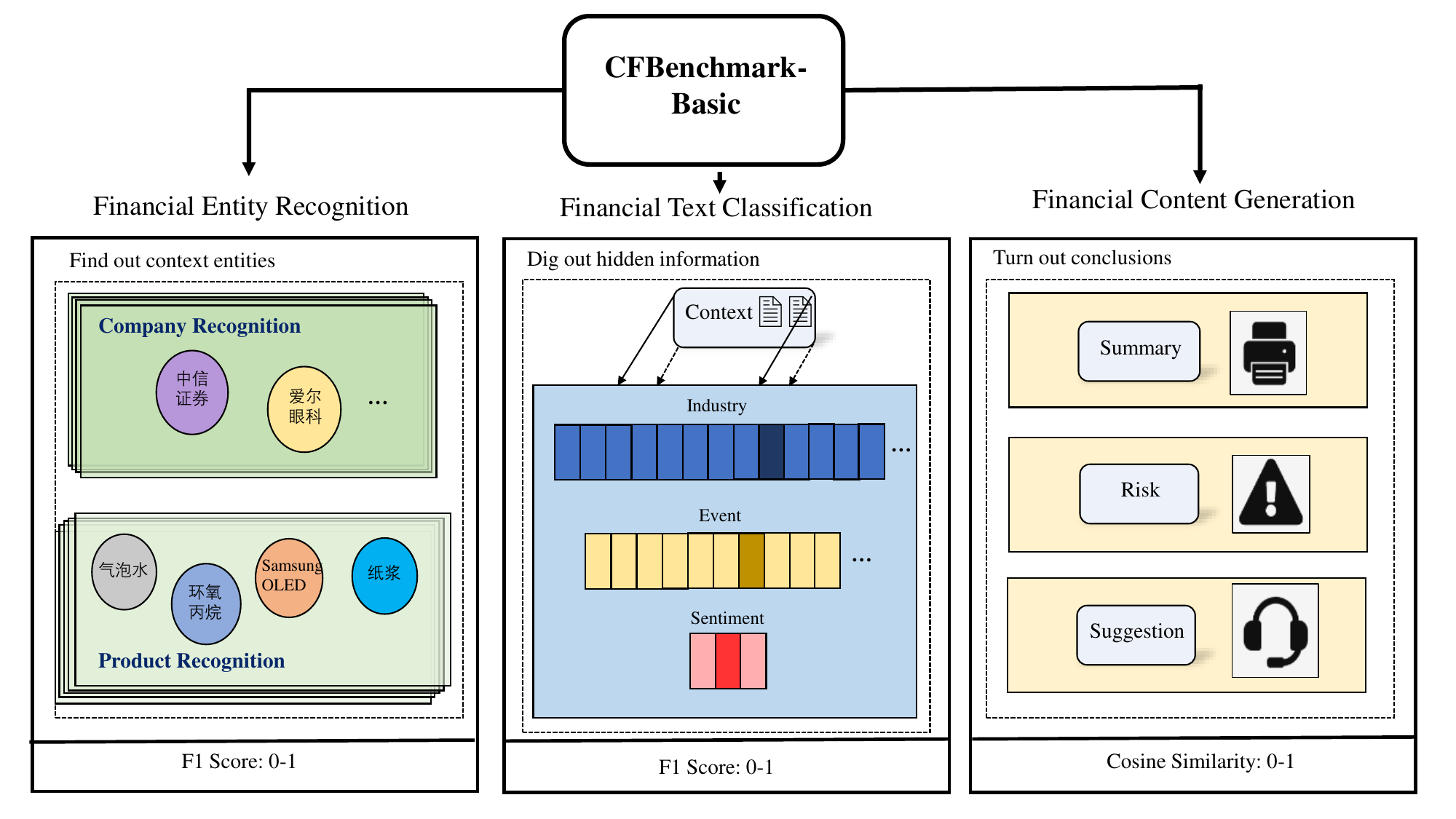}
   \caption{Overview of CFBenchmark-Basic, assessing systems from financial entity recognition, financial text classification, and financial content generation.}
   \label{fig:data_structure}
\end{figure*}

\section{CFBenchmark-Basic}

\subsection{Overview}
CFBenchmark-Basic is proposed to evaluate the financial text processing capabilities of LLMs in the financial domain from three aspects, financial entity recognition, financial text classification, and financial content generation, which are illustrated in Figure~\ref{fig:data_structure}.
To ensure the length and quality of the financial texts, we use financial news and research reports as the financial texts in our benchmark tasks
In financial entity recognition, we require LLMs to identify product and company names mentioned in the corresponding financial text.
For financial text classification, we task the LLMs with identifying the sector involved in the financial text, detecting the event in financial news~\cite{liang2021f} and analyzing the sentiment conveyed by the text~\cite{araci2019finbert}.
As for financial content generation, we expect the LLMs to generate the summary, risk alerts, and investment suggestions based on the research report.
Recognizing the entity such as companies and products can aid in classifying the industries, events, and sentiment in the financial content.
Classifying this information can further help summarize and refine the news content, generating abstract summaries, risk warnings, and advanced investment suggestions.
In fact, these three types of tasks are intrinsically linked to human cognitive processes.
By listing these three tasks in CFBenchmark-Basic, we aim to deconstruct this cognitive process.
By scoring LLMs separately in these three modules, we can comprehensively assess their capabilities in processing financial text.

\subsection{Task Definition}

\begin{figure*}[t]
  \centering
   \includegraphics[width=\textwidth]{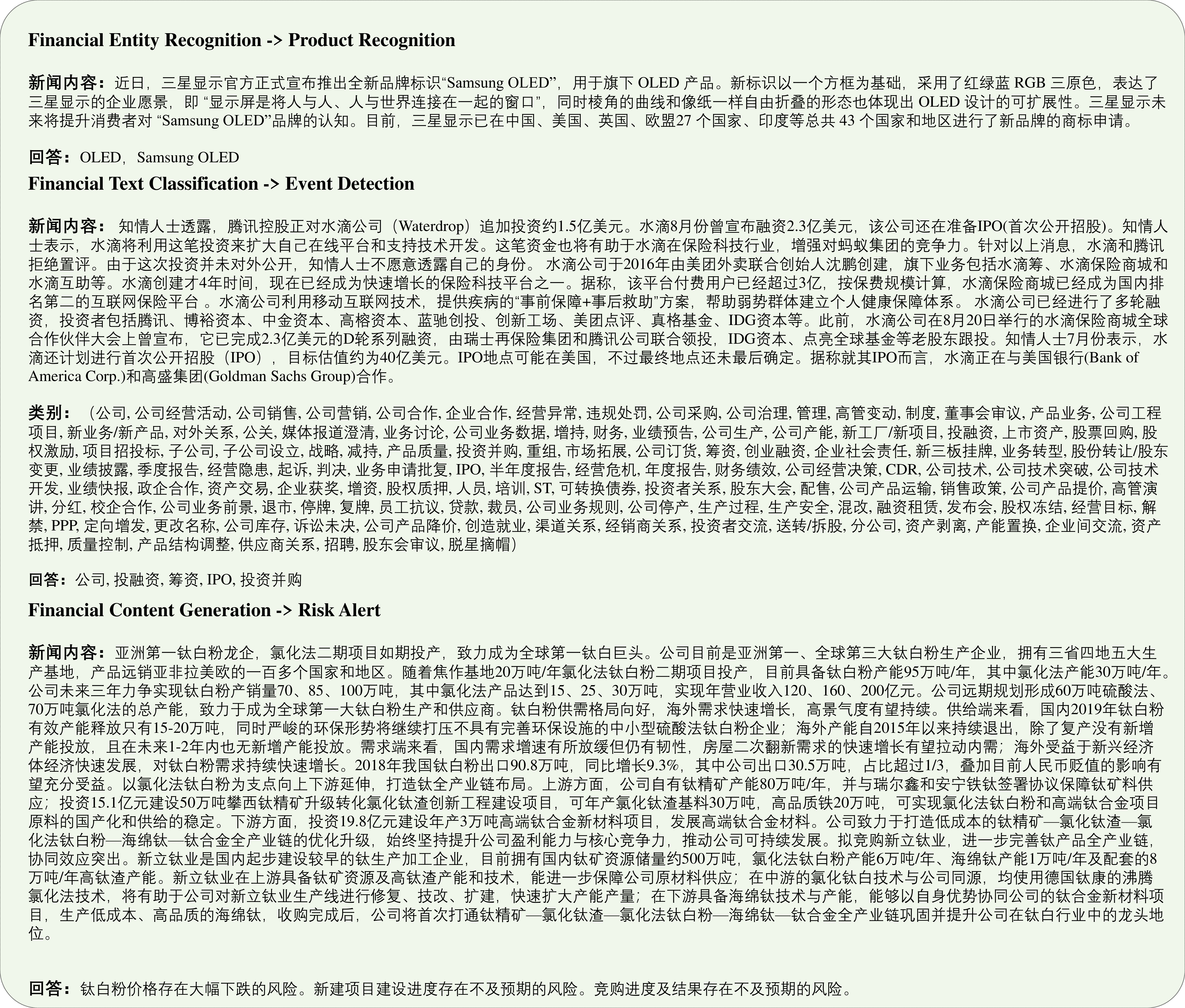}
   \caption{An example of our question context. For each aspect of financial entity recognition, financial text classification, and financial content generation, a representative case is presented.}
   \label{fig:data_example}
\end{figure*}

The CFBenchmark-Basic encompasses tasks from three aspects: financial entity recognition, financial text classification, and financial content generation, shown in Figure~\ref{fig:data_example}.
For financial entity recognition, we select two categories of entities to recognize, \emph{i.e.}, the companies and the products.
For financial text classification, we design three kinds of tasks, 1) to identify the sector of the financial text, 2) to detect financial events within the financial text, and 3) to analyze the sentiment of the financial text towards the market.
For financial content generation, report summaries, risk alerts, and investment suggestions were selected as the subjects of evaluation.
All financial texts are sourced from Chinese financial research reports and news data spanning from 2018 to 2023.

\subsubsection{Financial Entity Recognition}
In financial texts, ``product'' and ``company'' are two basic concepts, and usually play pivotal roles.
Companies are the primary active entities in the financial market, with the vast majority of economic and financial activities revolving around them.
Besides, products represent the operational domain of each company and serve as critical reference points for profit estimations and core competencies.
Therefore, the ability to correctly identify products and companies within a text is of paramount importance in the financial domain, which directly influences the assessment of the core content of news.
Thus, in financial entity recognition, we choose ``product'' and ``company'' as the target for identification.

\subsubsection{Financial Text Classification}\label{sec:data:task:2}

\textbf{Sector Identification:}
When addressing financial issues, senior analysts often start from a specific point and expand to a broader context, considering the macro backdrop of a given issue.
Applying this methodology to financial text processing implies that after identifying the products and companies within the news, determining their corresponding sector becomes imperative.
In standard sector classification\footnote{National Economic Industry Classification of China (GB/T4754-2022)}, it is designed as a hierarchical classification, including {29} first-level sector categories and {133} second-level sector categories.
The sector identification is defined as providing the LLMs with a first-level sector label and a set of second-level candidates, and then requiring them to select from these candidates.

\noindent\textbf{Event Detection:}
Apart from the industry category, financial events encompass a series of specific occurrences. These provide pivotal information and assist traders in making informed investment decisions in the financial market~\cite{liang2020f}. Detecting and understanding these events is essential to grasp the intricate relationships among various subjects within financial texts.
We follow the criterion in~\cite{liang2020f} to construct our event detection task, which includes 254 distinct events from four aspects, \emph{i.e.}, industry-related, macrography-related, market-related, and company-related.
Therefore, event detection is formed by providing the LLMs with aspect scope and a set of event candidates and then requiring them to detect the event from the given candidates.

\noindent\textbf{Sentiment Analysis:}
In traditional settings, sentiment analysis of the text reflects the attitude of the author.
However, the sentiment analysis in the financial domain is required to analyze how the financial text affects the sentiment in the financial market.
For example, news about reducing the staff usually represents a ``negative'' attitude in traditional sentiment analysis, however, in the financial domain, such news usually represents that a company reduces its human cost and increases its profit, which would bring a ``positive'' attitude to the market.
Such unique characteristics make sentiment analysis extremely important in financial text processing.
By combining sentiment with events and subjects, one can delve deeper into the underlying information embedded within the news.
For sentiment analysis, we adopt the most commonly used format: ``Positive-Neutral-Negative''.

\subsubsection{Financial Content Generation}
In financial content generation, we design three tasks that cover the basic requirement for LLMs, \emph{i.e.}, content summary, investment suggestions, and, risk alerts.
Content summary typically encompasses superficial information contained in the financial text, such as the main subjects and events, which evaluates the ability of LLMs to capture these surface-level details from the financial text.
Investment suggestions require LLMs to synthesize and infer from the gathered internal and external textual information, ultimately producing reasoned and concise advice.
Risk alerts are an essential component in financial assessments, which involves deep textual information mining related to the operational status, and investor sentiment towards a company, which is intrinsically related to the two preceding generation tasks and is essential to avoid potential risk.

\begin{table}[t]
\centering
\caption{The statistics of our CFBenchmark-Basic, which covers three aspects and eight tasks. We present the average text length and total number of questions for each task. }
\label{table:length}\vspace{5pt}
\setlength\tabcolsep{12pt}
\resizebox{\columnwidth}{!}{%
\begin{tabular}{llcc}
\hline\hline
Aspect                          & Task       & Avg-L        & \# Text \\ \hline
\multirow{2}{*}{Recognition}  & Company    & 512.63       & 273     \\
                                & Product    & 1643.30      & 297     \\ \hline
\multirow{3}{*}{Classification} & Sector   & 845.36       & 402     \\
                                & Event      & 661.38       & 577     \\
                                & Sentiment  & 617.87       & 591     \\ \hline
\multirow{3}{*}{Generation}     & Summary    & 959.85       & 593     \\
                                & Risk       & 945.91      & 591     \\
                                & Suggestion & 924.02       & 593     \\ \hline
-                               & Total      & - & 3917    \\ \hline\hline
\end{tabular}%
}
\end{table}

\subsection{Data Collection and Annotation}

To obtain the financial texts and annotations for our CFBenchmark-Basic, we follow~\cite{li2023cfgpt} and utilize a proxy-based distributed crawler to scrape relevant news and research report data.
Moreover, we follow steps in~\cite{zeng2021pangu,lu2023bbt} to filter and clean our benchmark data.
Regarding text annotations, we hired three professional researchers from financial institutions to annotate for the aforementioned eight financial tasks
We keep the financial text and corresponding answers that received consistent annotations from all three researchers in our CFBenchmark-Basic.

\begin{figure}[t]
  \centering
   \includegraphics[width=\columnwidth]{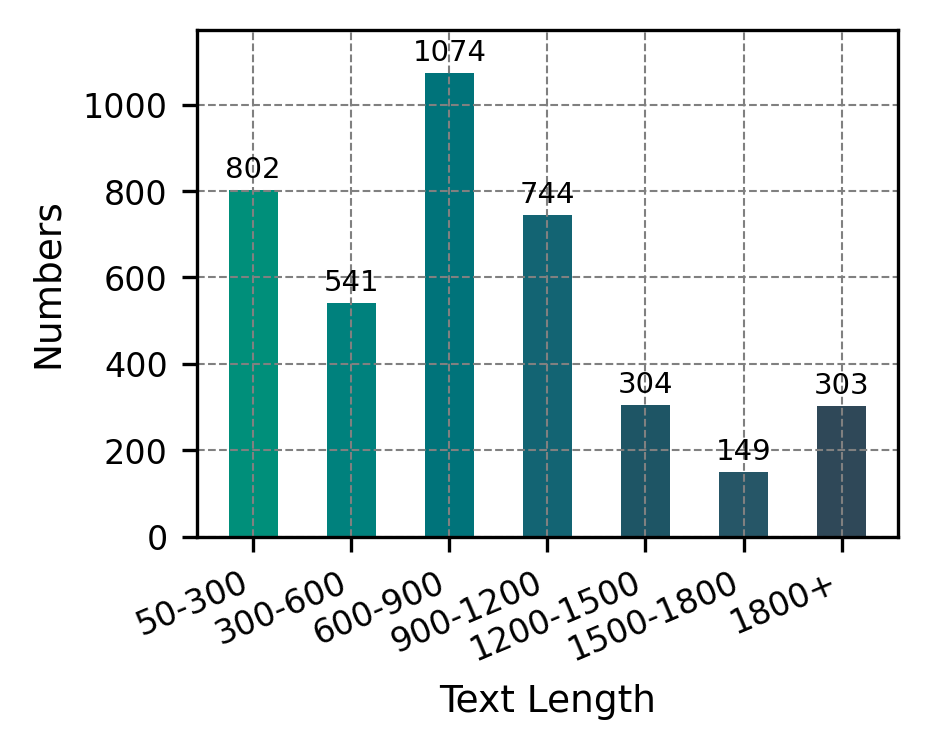}
   \caption{We illustrate the distribution of text lengths among these 3917 financial texts, where the length is calculated based on Chinese characters, starting from 50 and increasing in intervals of 300.}
   \label{fig:risk_assess}
\end{figure}

\subsection{Statistics}
In this section, we present the statistical information about our CFBenchmark-Basic.
Within our CFBenchmark-Basic, the financial texts are processed regarding three aspects, recognition, classification, and generation.
Specifically, the financial entity recognition task comprises 273 texts for company recognition and 297 texts for product recognition, where each financial text corresponds to 2.344 companies and 6.781 products on average.
The classification task includes 591 texts for sentiment analysis, 402 texts for sector identification, and 577 texts for event detection, where each financial text corresponds to 1.088 sector categories and 4.056 event categories on average.
The generation task consists of 593 texts for content summary, 591 texts for risk alerts, and 593 texts for investment suggestion, where the average length of responses of content summary, risk alerts, and investment suggestion are 76.258, 37.582, and 171.656, respectively.

Additionally, we also provide the statistical analysis on the lengths of financial texts, the details of which are presented in Table \ref{table:length}.
For financial entity recognition, the average length for company recognition is 512.63 characters, while for product recognition, the average length is 1643.39 characters.
For financial text classification, the average character lengths for sentiment analysis, sector identification, and event detection are 617.87, 845.35, and 661.38, respectively.
For financial content generation, these three tasks have the longest average text lengths, where the average character length for the risk alerts, content summary, and investment suggestions are 945.91, 959.85, and 924.02, respectively.
To our knowledge, the average input length for each of our tasks is longer than that of existing benchmarks.
When compared to other benchmarks, our benchmark has a longer average text length for each task, which is more closely aligned with real-world application scenarios~\cite{noever2023professional}.
We further illustrate the distribution of text lengths among these {3917} questions in Fig \ref{fig:risk_assess}.
The length is based on Chinese characters, starting from 50 and counting in intervals of 300. There are 802 items with a text length from 50 to 300, 541 items from 300 to 600, 1074 items from 600 to 900, 744 items from 900 to 1,200, 304 items from 1,200 to 1,500, 149 items from 1,500 to 1,800, and 303 items over 1,800.
The distribution demonstrates that our benchmark not only encompasses the evaluation of reading comprehension capabilities across various text lengths, which are considered the characteristics of the financial domain but also includes a significant number of longer texts that pose challenges for LLMs.
Moreover, the average lengths are in line with the length of news articles in reality.
Therefore, our benchmark exhibits a higher degree of practical accuracy.

\subsection{Evaluations}
In this section, we introduce the metrics for evaluating test results.
We utilized two types of metrics to evaluate the performance of LLMs in the financial domain on our CFBenchmark-Basic.
For recognition and classification tasks, we employed the F$_1$-score as the evaluation metric, which balances the precision and recall.
For the generation tasks, we utilize cosine similarity between the vector representation of ground truth and the generated answer to measure the generation ability.
Since there are usually different expressions with similar meanings in our generation tasks, simply employing Rough-Score or BLEU-score is not reasonable.
Specifically, the bge-zh-v1.5~\cite{bge_embedding} serves as the oracle model to generate the sentence embedding.
We calculate evaluation scores for each sub-task individually and provide the average score for each category.

\begin{table*}[!tbp]
\setlength\tabcolsep{3pt}
\caption{The results on CFBenchmark-Basic on renowned and representative models.
We present the results of running our proposed benchmark tests on 22 LLMs.
For each category in the three types of tasks, we computed the average value across the sub-tasks to represent the capability of LLM in the corresponding aspect.
For generation, we used cosine similarity as the metric.
For recognition and classification, we employed the F$_1$-score. The best values in each column are highlighted in bold.
}
\label{table:result}
\vspace{5pt}
\centering
\resizebox{\textwidth}{!}{%
\begin{tabular}{l|c|cc|c|ccc|c|ccc|c|c}
\hline\hline
 Model & Size&  Company &  Product &  R.Avg & Sector & Event & Sentiment & C.Avg & Summary & Risk & Suggestion & G.Avg & Avg\\
\hline\hline
Human              & -   &0.931 & 0.744 & 0.838 & 0.975 & 0.939 & 0.912 & 0.942 & 1.000 & 1.000 & 1.000 & 1.000 & 0.927\\
\hline
ChatGPT            & 175B& 0.797 & 0.198 & 0.498 & 0.453 & 0.458 & 0.425 & 0.455 & 0.593 & 0.541 & 0.771 & 0.635 & 0.529 \\
ERNIE-Bot          & 260B& 0.807 & 0.300 & 0.533 & 0.408 & 0.350 & 0.186 & 0.315 & 0.715 & 0.590 & 0.716 & 0.673 & 0.507 \\
ERNIE-Bot-4        & -   & 0.819 & 0.417 & 0.618 & 0.418 & 0.358 & 0.375 & 0.384 & 0.721 & 0.629 & 0.718 & 0.689 & 0.564 \\
Falcon-7b          & 7B  & 0.671 & 0.168 & 0.420 & 0.169 & 0.132 & 0.250 & 0.184 & 0.302 & 0.301 & 0.246 & 0.283 & 0.296 \\
Falcon-7b-chat     & 7B  & 0.582 & 0.046 & 0.314 & 0.112 & 0.142 & 0.153 & 0.135 & 0.307 & 0.299 & 0.258 & 0.288 & 0.246 \\
bloomz-7b1         & 7B  & 0.765 & 0.166 & 0.465 & 0.252 & 0.154 & 0.394 & 0.267 & 0.451 & 0.371 & 0.462 & 0.428 & 0.387 \\
bloomz-7bt1-mt     & 7B  & 0.751 & 0.157 & 0.454 & 0.087 & 0.182 & 0.380 & 0.216 & 0.425 & 0.379 & 0.396 & 0.400 & 0.357 \\
Qwen-7B            & 7B  & 0.780 & 0.357 & 0.569 & 0.480 & 0.335 & 0.379 & 0.398 & 0.750 & 0.505 & 0.713 & 0.656 & 0.541 \\
Qwen-Chat-7B       & 7B  & 0.763 & 0.360 & 0.562 & 0.400 & 0.367 & 0.265 & 0.344 & 0.548 & 0.307 & 0.379 & 0.411 & 0.439 \\
Qwen-14B           & 14B & 0.805 & 0.421 & 0.613 & 0.481 & 0.350 & 0.385 & 0.405 & 0.754 & 0.608 & 0.717 & 0.693 & 0.570 \\
Qwen-Chat-14B      & 14B & 0.814 & 0.442 & 0.628 & 0.382 & 0.400 & 0.350 & 0.377 & 0.732 & 0.478 & 0.736 & 0.649 & 0.551 \\
ChatGLM2-6B        & 6B  & 0.747 & 0.313 & 0.530 & 0.285 & 0.300 & 0.357 & 0.314 & 0.657 & 0.454 & 0.671 & 0.594 & 0.479 \\
Baichuan2-7B-Base  & 7B  & 0.672 & 0.340 & 0.506 & 0.342 & 0.490 & 0.480 & 0.437 & 0.739 & 0.619 & 0.751 & 0.703 & 0.549 \\
Baichuan2-7B-Chat  & 7B  & 0.757 & 0.402 & 0.579 & 0.425 & 0.475 & 0.323 & 0.408 & 0.725 & 0.648 & 0.732 & 0.702 & 0.563 \\
Baichuan2-13B-Base & 13B & 0.781 & 0.330 & 0.555 & 0.436 & 0.496 & 0.477 & 0.470 & 0.725 & 0.503 & 0.747 & 0.658 & 0.561 \\
Baichuan2-13B-Chat & 13B & 0.797 & 0.314 & 0.556 & 0.472 & 0.507 & 0.387 & 0.455 & 0.739 & 0.634 & 0.746 & 0.706 & 0.572 \\
InternLM-7B        & 7B  & 0.612 & 0.233 & 0.423 & 0.266 & 0.311 & 0.328 & 0.302 & 0.378 & 0.336 & 0.379 & 0.364 & 0.363 \\
InternLM-7B-Chat   & 7B  & 0.632 & 0.261 & 0.447 & 0.272 & 0.364 & 0.399 & 0.345 & 0.363 & 0.270 & 0.353 & 0.329 & 0.374 \\
InternLM-20B       & 20B & 0.809 & 0.358 & 0.583 & 0.500 & 0.427 & 0.417 & 0.448 & 0.706 & 0.653 & 0.728 & 0.695 & 0.575 \\
InternLM-20B-Chat  & 20B & 0.488 & 0.362 & 0.425 & 0.323 & 0.327 & 0.370 & 0.340 & 0.706 & 0.578 & 0.762 & 0.662 & 0.476 \\
\hline
CFGPT1-stf-LoRA     & 7B  & 0.820 & 0.414 & 0.617 & 0.569 & 0.729 & 0.769 & 0.689 & 0.745 & 0.584 & 0.609 & 0.646  & 0.650 \\
CFGPT1-sft-Full     & 7B  & \textbf{0.836} & \textbf{0.476} & \textbf{0.656} & \textbf{0.700} & \textbf{0.808} & \textbf{0.829} & \textbf{0.779} & \textbf{0.798} & \textbf{0.669} & \textbf{0.808} & \textbf{0.758}  & \textbf{0.731} \\
\hline\hline
\end{tabular}
}

\end{table*}

\section{Experiments}
In this section, we first present our experiment setting~(Section~\ref{sec:exp:setup}), evaluated methods~(Section~\ref{sec:exp:method}), followed by results and discussions~(Section~\ref{sec:exp:result}).

\subsection{Experiment Setup}~\label{sec:exp:setup}
In this section, we detail our experimental setup based on the financial domain LLMs benchmarks.
To evaluate the adaptability and enhance the performance of the model, we employed two evaluation methods.
Additionally, we will provide examples to illustrate our approach to constructing prompts.

\noindent\textbf{Zero-shot}~\cite{kojima2205large}:
In the zero-shot setting, the model directly receives instructions and corresponding text without exposure to similar test instances.
This scenario aims to assess the inherent capability of LLMs to tackle financial problems without relying on external cues.
For the model input, both the prompt and financial text are directly fed into the model.

\noindent\textbf{Few-shot}~\cite{brown2020language}:
In contrast to zero-shot, the few-shot evaluation specifically assesses the potential of LLMs in handling financial tasks.
For each type of question in this evaluation, we create a set of sample inputs for the model.
Unlike some existing financial domain LLM benchmarks, considering the input length limitation of the models, we only choose three samples as the example inputs in the few-shot scenario.
For the model input, the prompt and financial text are fed into the model along with the samples.

\subsection{Models}~\label{sec:exp:method}
To obtain a comprehensive and in-depth evaluation of large models in processing Chinese financial texts, we selected 22 representative and outstanding LLMs for testing.

\noindent\textbf{GPT}~\cite{chatgpt,gpt4}: ChatGPT, also known as GPT-3.5-TURBO, is a widely-known LLM in the NLP domain. It is distinguished for its cost-efficiency and versatile functionalities, finding utility in a diverse array of applications.

\noindent\textbf{ERNIE-Bot}~\cite{sun2021ernie}: ERNIE-Bot and ERNIE-Bot-4 are introduced by Baidu-Inc. These models incorporate principles from both deep learning and knowledge augmentation domains during their development.

\noindent\textbf{Qwen}~\cite{bai2023qwen}: Qwen-7B and Qwen-14B are LLMs introduced by Alibaba Cloud. Qwen-Chat-7B and Qwen-Chat-14B are tailored versions of Qwen-7B and Qwen-14B, respectively. Their primary training objective is to function as adept AI assistants.

\noindent\textbf{Baichuan}~\cite{yang2023baichuan}: Baichuan-7B-Base is based on the transformer architecture, introduced by Baichuan-inc. It is adept at handling both Chinese and English NLP tasks. Baichuan-7B-Chat is the refined version, tailored from the pre-trained Baichuan-7B-Base model. In a similar vein, its 13B parameter counterparts, specifically Baichuan-13B-Base and Baichuan-13B-Chat, are also chosen for evaluation.

\noindent\textbf{ChatGLM}~\cite{du2021glm,zeng2022glm}: ChatGLM2-6B is an expansive generative model, honed on a bilingual corpus encompassing both Chinese and English. It is predicated on the General Language Model framework, known as GLM-6B. Notably, this model signifies the enhanced successor of the original ChatGLM-6B.

\noindent\textbf{BLOOM}~\cite{muennighoff2022crosslingual,scao2022bloom}: Bloomz-7b1-mt is a multilingual model grounded in the BLOOM framework. Its capabilities have been further refined and augmented through a sophisticated process of multitasking prompting, leading to enhanced performance across diverse tasks and languages.

\noindent\textbf{Falcon}~\cite{almazrouei2023falcon}: Falcon-7B is a large language model devised by TII. Unique in its architecture, it employs a causal, decoder-only structure. Integrating advanced features such as FlashAttention~\cite{Dao2022flashattention} and multi-query attention mechanisms from~\cite{Noam2019fast}, Falcon-7B is optimized for high-performance tasks in natural language processing. Considering the size of this model, we evaluate the 7B version.

\noindent\textbf{InternLM}~\cite{team2023internlm}: InternLM-7B and InternLM-20B are large-scale generative models that could possess the capability to handle texts within a maximum length of 4096.


\noindent\textbf{CFGPT:} CFGPT~\cite{li2023cfgpt} is a large language model for Chinese financial assistant. To enhance the capability, CFGPT further pre-trains the InternLM-7b on the financial documents, including announcements, research reports, social media content, and financial news articles, and get CFGPT1-pt-7B. Then CFGPT fine-tune the pre-trained model on six financial tasks, including sentiment analysis, event detection, report summarization, topic decomposition, question answering, and stock movement prediction with LoRA and full-parameters, and get the models named: CFGPT1-sft-7B-LoRA and CFGPT1-sft-7B-Full. Note that the financial texts in the CFBenchmark are not utilized for further pre-training or supervised finetuning of the CFGPT.

\subsection{Results}~\label{sec:exp:result}
In this section, we conduct an in-depth assessment of the selected LLMs on our CFBenchmark-Basic.
Table \ref{table:result} delineates the performance metrics for each respective task and subsequently consolidates these results for every task category.
This detailed representation underscores the proficiency and specialization of the LLMs across diverse domains.

In the domain of financial entity recognition, {Qwen-Chat-14B} emerges as a frontrunner, registering an F$_1$-Score of {0.628}.
It is closely trailed by {ERNIE-Bot-4} and {Qwent-14B}, who clock in F$_1$-Scores of {0.618} and {0.613}, respectively. Noteworthily, all models showcased superior performance in identifying companies over products.
This can be attributed to the fact that financial texts typically feature a broader spectrum of products than companies.
{ChatGPT}, in particular, exhibited a reduced efficacy in product identification, largely due to its overarching focus on company categorization.
This inclination was observed even when presented with few-shot examples, a trend mirrored across other models.
While achieving a commendable score in company recognition is laudable, the suboptimal performance in product recognition underscores the potential for refinement and enhancement in the ability of LLMs to comprehend and differentiate multiple subjects within text.
The performance of LLMs under this segment exhibit two primary deficiencies.
Firstly, they tend to omit entities that require to be recognized.
Secondly, they erroneously classify non-relevant entities within the text as pertinent.
For instance, the models indiscriminately incorporate products, personal names, company and sector designations into the product recognition responses.

In the realm of financial text classification, {Baichuan2-13B-Base} outshines its peers, registering a score of {0.470}.
This is closely followed by {Baichuan2-13B-Chat} and {InternLM-20B}, achieving scores of {0.455} and {0.448}, respectively.
Given this scenario as a 3-label classification task, the efficacy of the current LLMs in sentiment analysis is notably subpar.
As articulated in Section~\ref{sec:data:task:2}, sentiment analysis within the financial sphere substantially deviates from its general domain counterpart, accounting for the observed performance deficit.
Evaluating the scores across sector identification and event detection tasks further reiterates the existing gap.
It underscores the considerable opportunity for refining and advancing the capabilities of the current LLMs.
In this segment, the primary issue with the responses from LLMs is their inaccurate categorization, indicating a superficial understanding of the concepts to be classified. Furthermore, these models often generate answers that fall outside the required categories, even when a few-shot evaluation approach is introduced.

In the dimension of financial content generation, {Baichuan2-13B-Chat} takes the lead, registering a score of {0.706}. The most competitive followers are {Baichuan2-7B-Base} and {Baichuan2-7B-Chat}, tallying scores of {0.703} and {0.702}, respectively.
Delving deeper into the triad of sub-tasks in this category, a recurring trend emerges: models tend to lag in risk alert generation compared to their prowess in crafting content summaries and investment suggestions.
This underscores the nuanced demands of risk alerts, encapsulating both internal and external knowledge domains.
Sole dependence on financial texts proves insufficient for this complicated task.
A juxtaposition of performances across these sub-tasks reveals an interesting observation: suboptimal results in financial entity recognition and text classification do not drastically impede content generation prowess.
This suggests that the generative strengths of current LLMs might not be rooted in their reasoning capabilities. Consequently, infusing genuine reasoning into text generation remains a pivotal challenge within the financial domain.
In this segment of tasks, although the LLMs score relatively higher on average compared to the other two segments, a closer look at their responses reveals that the content is rather generic and repetitive. For instance, in risk alert, where the average scores are lower, each category of LLMs tends to produce replies mentioning ``industry competition risk'', ``macroeconomic downturn risk'', and the similar phrases repetitively, demonstrating a lack of specificity in the generated content with respect to the text.

Upon comparing the average scores across various tasks, we observe that multilingual LLMs, such as the Falcon and BLOOMZ series, performed less favorably than bilingual Chinese-English models of comparable parameter size.
Multilingual models exhibited weaker capabilities in capturing information from Chinese financial texts compared to those trained exclusively on Chinese corpus, which effectively underscores the regional specificity of financial problems as revealed by our benchmark.
On the other hand, LLMs typically outperform their smaller counterparts within the same model series.
This reflects that models with a greater number of parameters have a stronger ability to extract information from texts.
This is particularly aligned with the characteristics of the financial sector, which always demands the precise extraction of pertinent information from vast datasets.

In the comparison of CFGPT1-sft-7B-LoRA and CFGPT1-sft-7B-Full, we observe a notable superiority of the CFGPT1-sft-7B-LoRA, especially when matched against LLMs with a comparable parameter count in the 6B to 7B range. The CFGPT1-sft-7B-Full reigns supreme over all baseline methods within our CFBenchmark-Basic. This suggests the important role of domain-specific modeling in enhancing performance outcomes.

\section{Conclusion}
In this work, we analyzed the requirements of LLMs in the financial domain, based on which we provided the roadmap for an LLM benchmark system in the Chinese financial domain.
Besides, we introduced a basic version of the Chinese financial benchmark, CFBenchmark-Basic, which covers three aspects and eight tasks, including financial entity recognition of company and product, financial text classification about sentiment, sector, and event, and financial content generation about content summary, risk alert, and investment suggestion.
Regarding the source of information, CFBenchmark-Basic has encompassed a series of high-quality Chinese textual data and annotations, making the evaluation more aligned with real-world financial application scenarios.
Moreover, we conducted extensive experiments to evaluate the 22 LLMs available in the literature, and revealed that there is still significant room to improve LLMs in basic tasks of financial text processing.
In the future, we will continue to contribute more comprehensive tasks, the advanced version of CFBenchmark, to evaluate the performance of LLMs for Chinese financial assistant.

\bibliography{anthology,custom}
\bibliographystyle{acl_natbib}



\clearpage

\appendix

\appendix
\section{Appendix}
\subsection{Zero-shot example}
An instance of zero-shot evaluation in the CFBenchmark-Basic.

\begin{figure}[h]
    \centering
    \includegraphics[width=1.99\linewidth]{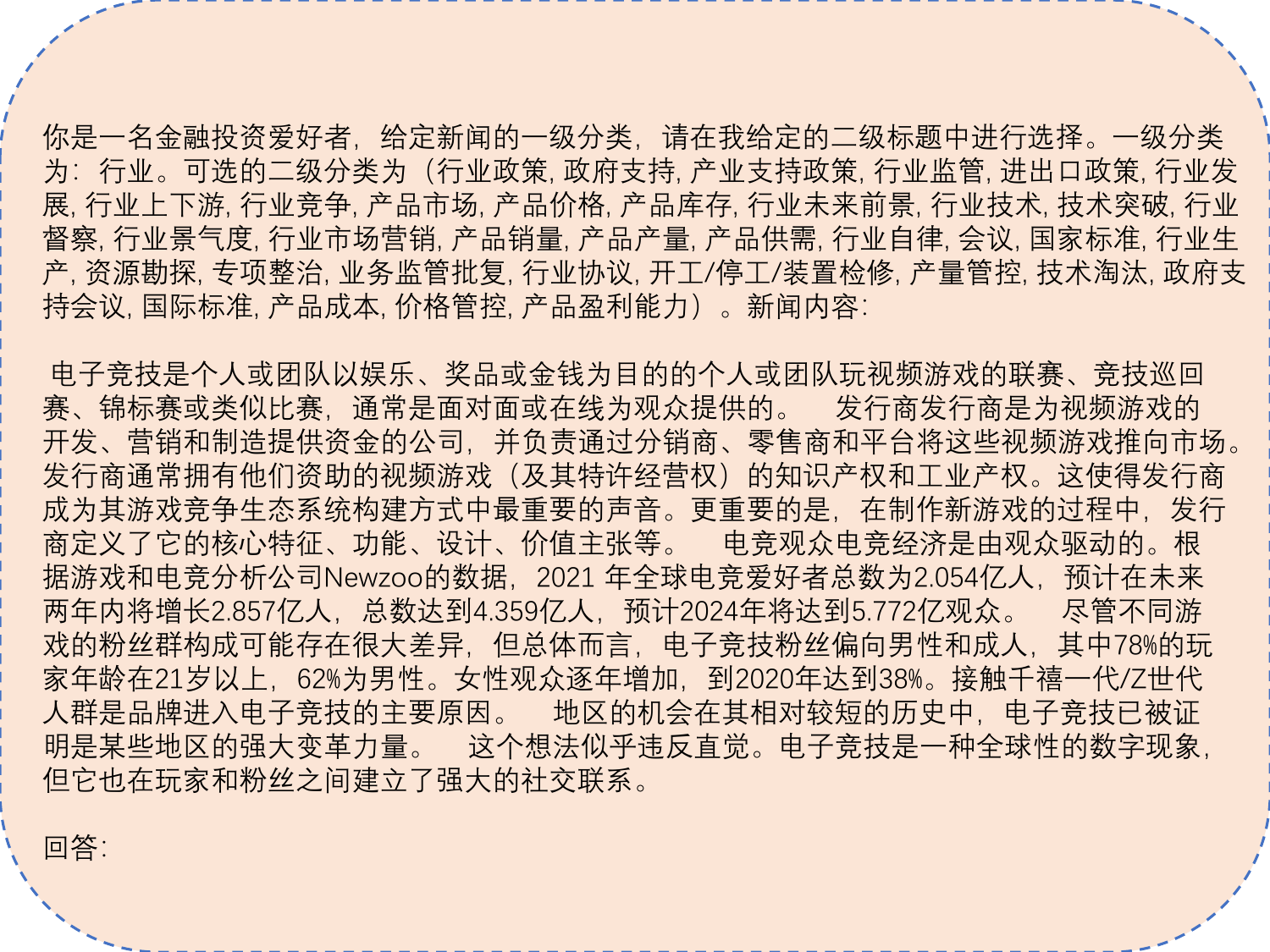}
    \label{fig:zeroshot}
\end{figure}

\clearpage
\subsection{Few-shot example}
An instance of few-shot evaluation in the CFBenchmark-Basic.
\begin{figure}[h]
    \centering
    \includegraphics[width=1.99\linewidth]{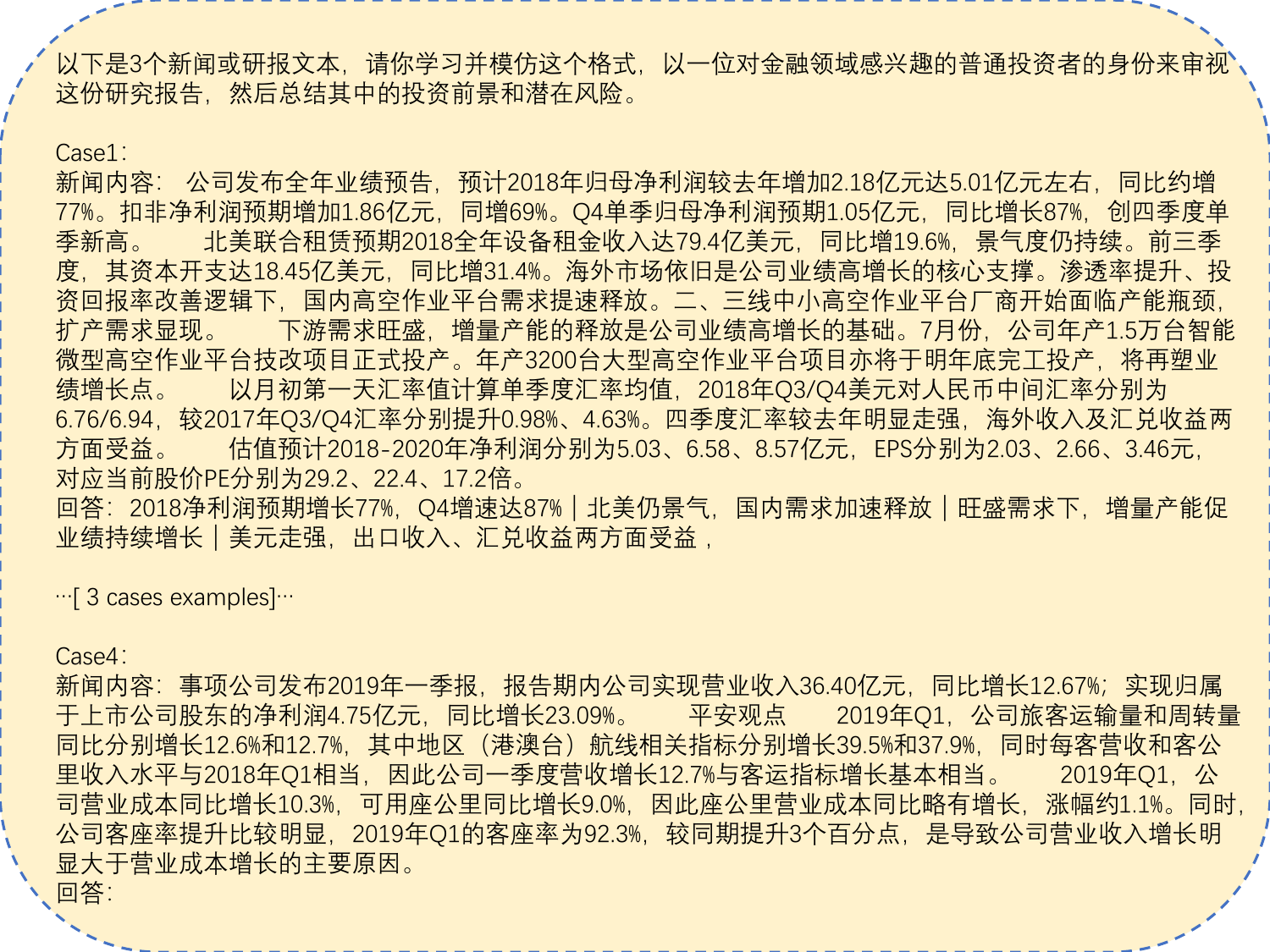}
    \label{fig:fewshot}
\end{figure}
\end{document}